\pdfoutput=1

\documentclass[11pt]{article}

\usepackage[preprint]{acl}

\usepackage{times}
\usepackage{latexsym}

\usepackage[T1]{fontenc}

\usepackage[utf8]{inputenc}

\usepackage{microtype}

\usepackage{inconsolata}

\usepackage{graphicx}
\usepackage{boxedminipage}
\usepackage{amsmath}
\usepackage{multirow}
\usepackage{arydshln}
\usepackage{tikz}
\usepackage{pgfplots}
\usepackage{graphicx,subfigure}

\usepackage{float}
\title{New Evaluation Paradigm for Lexical Simplification}

\author{
	Jipeng Qiang$^{1}$,Minjiang Huang$^{1}$,Yi Zhu$^{1}$, Yunhao Yuan$^{1}$, Chaowei Zhang$^{1}$,Xiaoye Ouyang$^{2}$ \\
	$^1$ School of Information and Engineering, Yangzhou University, Jiangsu, China\\
	$^2$ China Academy of Electronic and Information Technology, Beijing, China\\
    \texttt{\{jpqiang, zhuyi, yhyuan, cwzhang\}@yzu.edu.cn}, \texttt{9624219@qq.com}\\ \texttt{ouyangxiaoye@cetc.com.cn}
}

\begin{document}
\maketitle
\begin{abstract}
Lexical Simplification (LS) methods use a three-step pipeline: complex word identification, substitute generation, and substitute ranking, each with separate evaluation datasets. We found large language models (LLMs) can simplify sentences directly with a single prompt, bypassing the traditional pipeline. However, existing LS datasets are not suitable for evaluating these LLM-generated simplified sentences, as they focus on providing substitutes for single complex words without identifying all complex words in a sentence.

To address this gap, we propose a new annotation method for constructing an all-in-one LS dataset through human-machine collaboration. Automated methods generate a pool of potential substitutes, which human annotators then assess, suggesting additional alternatives as needed. Additionally, we explore LLM-based methods with single prompts, in-context learning, and chain-of-thought techniques. We introduce a multi-LLMs collaboration approach to simulate each step of the LS task. Experimental results demonstrate that LS based on multi-LLMs approaches significantly outperforms existing baselines.
\end{abstract}

\section{Introduction}

The aim of Lexical Simplification (LS) is to substitute complex words within sentences with simpler alternatives, thereby enhancing reading comprehension for a diverse array of readers, including non-native speakers \citep{paetzold2016unsupervised} and individuals with cognitive impairments \citep{saggion2017automatic}. LS method is commonly framed as a pipeline of three main steps: Complex Word Identification (CWI), Substitute Generation (SG), and Substitute Ranking (SR) \citep{paetzold2017survey,qiang2021lsbert}.  The three steps are treated as different independent tasks, and they have their own evaluation datasets.  

The CWI step identifies complex words or phrases in a text, which can be treated as classification or sequence labeling task \cite{yimam2018report,paetzold2017lexical,gooding2019complex,qiang2021lsbert}. The SG step generates many simpler alternatives for complex words by Linguistic-based methods \cite{pavlick2016simple,maddela-xu-2018-word}, word embedding-based methods \cite{glavavs2015simplifying,paetzold2017lexical}, pretrained language model-based methods \cite{qiang2020AAAI,LiuQLYZH23}, or Large language model (LLM) based methods \cite{sheang2023multilingual}. The SR step ranks the generated substitutes of the SG step by considering the context and the complex word \cite{paetzold2017lexical,qiang2021chinese}. This step typically combines multiple features for ranking, including word frequency, meaning preserved, contextual fluency, and so on. 

\begin{figure}
  \centering
   \includegraphics[width=78mm]{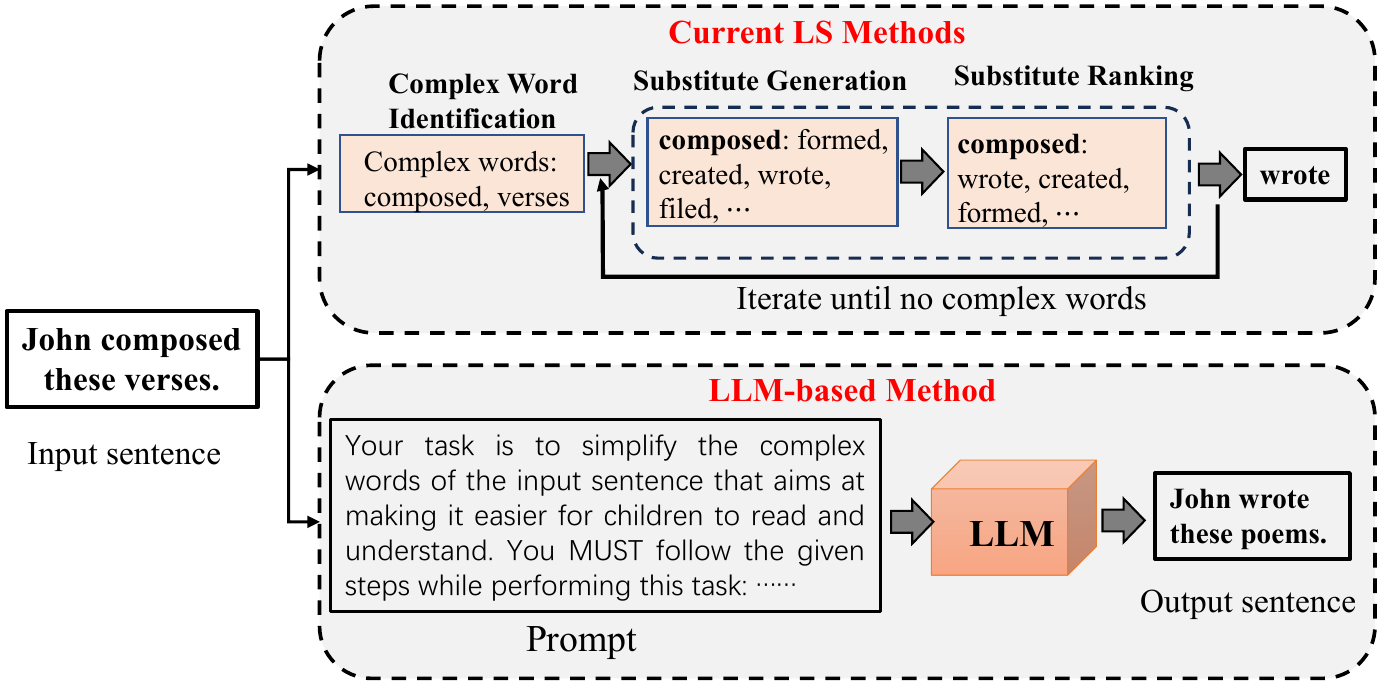} \\
 \caption{Flowchart of current LS methods and LLM-based LS method. Current LS methods require not only three models for three steps: complex word identification, substitute generation, and substitute ranking, but also iterative simplification of each complex word. We found that the LLM-based method only needs a single prompting to complete the task.}
 \label{LLMFig1}
\end{figure}

Recently, we have found that LLM can directly output simplified sentences that satisfy the LS requirements as long as the appropriate prompt template is designed. As shown in Figure~\ref{LLMFig1}, after feeding the original sentence "John composed these verses" into LLMs,   two simplified sentences generated from LLMs (GPT-3.5and GPT-4o-mini) ("John wrote these poems" and "John composed these poems"), which one is better?. LLM-based method shows its advantage without a pipeline of three steps. However, it brings a new problem: Can the existing LS datasets evaluate this type of simplified sentence?

CWI datasets \citep{yimam2018report,gooding2018camb} only provide all the complex words for each sentence.  Each instance in LS datasets \cite{paetzold2017lexical,qiang2021lsbert} consists of a sentence, a complex word, and a list of gold candidates. These datasets only focus on providing many substitutes for one complex word and do not identify all complex words within a sentence. We cannot use the existing datasets to evaluate the LS performance of different LLMs.

The evaluation of such methods should consider the following two aspects: (1) Methods that correctly simplify a greater number of complex words should exhibit better performance. (2) Considering that the difficulty of complex words varies, methods that simplify more challenging complex words should demonstrate superior performance. Constructing a dataset that can directly evaluate this type of simplification will open up a new evaluation paradigm for LS task. 

In this paper, we will propose a novel annotation method to construct an all-in-one style LS Dataset based on human-computer collaboration. Considering that existing CWI datasets have already annotated complex words, we built LS datasets based on the CWI datasets, thus saving the workload of annotating complex words. Given a complex word, we did not manually annotate the substitutes; instead, we used a human-computer interaction approach to generate the substitutes, since multiple automated LS methods can swiftly generate a vast pool of potential substitutes, alleviating the burden on human annotators. Subsequently, human annotators assess the suitability of these alternatives.

To address the LS task, we explore not only LLM-based methods with a single prompt but also integrate in-context learning and chain-of-thought techniques. To further improve the performance of LLM-based methods, we propose a novel method LLM via multi-LLMs collaboration.  This work utilizes multi-LLM collaboration to simulate each step (CWI, SG, and SR) of LS task. Experimental results show that LLM-based methods outperform existing multi-step approaches, and the multi-LLMs-based method significantly outperforms LLM-based method. 

\section{Related Work}
All existing LS systems are framed as a pipeline of three or four steps: Complex Word Identification, Substitute Generation, Substitute Selection (SS, optional), and Substitute Ranking (SR) \cite{paetzold2017survey,qiang2021lsbert}. We will cover the following three areas (CWI, SG, and SR).

\textbf{CWI.}  Early methods \cite{paetzold2017survey} used manual features and traditional machine learning. Later, the research included ensemble-based classifier methods, and the field expanded to multilingual CWI with notable milestones like CWI 2018 shared task \cite{yimam2018report,gooding2018camb}. The rise of deep learning introduced neural network models that outperformed earlier techniques, particularly with the use of word embeddings. Recent advances, such as BERT-based models \cite{qiang2021lsbert}, further improved performance by capturing context-dependent meanings. Hybrid models combining various methods have also shown promise \cite{north2024multils}. Standardized datasets and evaluation metrics, such as those from the Lexical Complexity Prediction 2021 Shared Task \cite{shardlow-etal-2021-semeval}, have been crucial for progress. The field continues to evolve, aiming to create more robust cross-lingual models and adaptive systems, enhancing text accessibility and readability for diverse applications \cite{north2023lexical}. 

\textbf{SG.} Early methods, e.g., linguistic-based methods \cite{pavlick2016simple,maddela-xu-2018-word} and word embedding-based Methods \cite{glavavs2015simplifying,paetzold2017lexical}, leverage linguistic resources or word embedding models to generate simpler substitutes. 

Pretrained language models have significantly advanced the field of lexical simplification \cite{sheang-etal-2022-controllable}. Qiang et al. \cite{qiang2020AAAI} and Liu et al. \cite{LiuQLYZH23} adopted pretrained language models to generate context-aware substitutes. These approaches leverage the contextual understanding of PLMs, resulting in more accurate and contextually appropriate simplifications.  The advent of large language models (LLMs) such as GPT-3 and ChatGPT has opened new possibilities for substitute generation. Most of the methods have shifted to exploring simpler approaches based on prompt engineering \cite{aumiller-gertz-2022-unihd}. Seneviratne et al. \cite{seneviratne-suominen-2024-anu} employed three prompt templates alongside an ensemble approach for SG. 

\textbf{SR.} Linguistic-based methods \cite{wubben2012,bott2012Can} leverage linguistic resources and predefined rules to rank substitutes. These methods often use frequency counts from corpora, word length, and predefined simplicity metrics to determine the rank of each substitute. Recent methods \cite{qiang2020AAAI, LiuQLYZH23} employed BERTScore or BARTScore to generate contextual embeddings and rank substitutes by considering both the local context and the global semantics. These methods benefit from the ability to capture complex contextual dependencies and produce more accurate rankings. 

Existing LLM techniques have only been applied to the SG step in LS task. In this paper, we will explore how to evaluate performance when a large model can solve the LS task in a single step. We will also investigate how multi-LLM techniques can further enhance the performance of large models.

\section{Creating Dataset}

In this section, we describe our human-machine collaboration method to build an all-in-one LS dataset. 
 
\subsection{Limitation of Existing LS dataset}

\begin{table}
\centering
\begin{tabular}{l|l}
\hline

\multirow{3}{*}{CWI} &  Guatemalan(1) Supreme Court  \\ 
&approves(1) impeachment(18) of Pre- \\
& sident Molina Yesterday  in Guatemala.\\ \cdashline{1-2} 

\multirow{5}{*}{LS}  & A man and a woman questioned on \\
& suspicion of assisting an offender \\
& {[}criminal; wrongdoer; convict; culprit;  \\
& violator; felon; accused{]}  have been \\
& released.\\ \cdashline{1-2} 

\multirow{5}{*}{Ours}  & Guatemalan Supreme Court \\
&approves(2) {[}allows; agrees; accepts;  \\
&passes{]} impeachment(18){[}removal; \\
&  dismissal{]} of President Molina \\
& Yesterday in Guatemala. \\
\hline
\end{tabular}
\caption{Comparison of different types of LS datasets. The CWI dataset annotates all complex words in each sentence, regardless of whether there are suitable substitutes for these complex words, such as proper nouns. The existing LS dataset provides only the substitutes for a complex word in each sentence, even though the sentence may contain many complex words. Our dataset provides all complex words that can be simplified and offers a corresponding set of simple words for each complex word. The number in parentheses indicates how many out of 20 annotators consider the word to be complex. Words in square brackets are given as simplified words.}
\label{tab:samples}
\end{table}

Existing CWI datasets in CWI 2018 shared task \citep{yimam2018report,gooding2018camb} provide all the complex words for each sentence, with each sentence annotated by 10 native and 10 non-native speakers with words that they found to be hard to children or non-native speakers to understand. Current LS datasets contain instances composed of a sentence, a target complex word, and a set of suitable substitutes  \citep{paetzold2017survey,aumiller-gertz-2022-unihd}. When LLMs can directly output a simplified sentence that meets the LS requirements, existing CWI or LS datasets have notable limitations that hinder their effectiveness in evaluating simplification models, as shown in Table \ref{tab:samples}. 

\begin{figure}
  \centering
   \includegraphics[width = \linewidth]{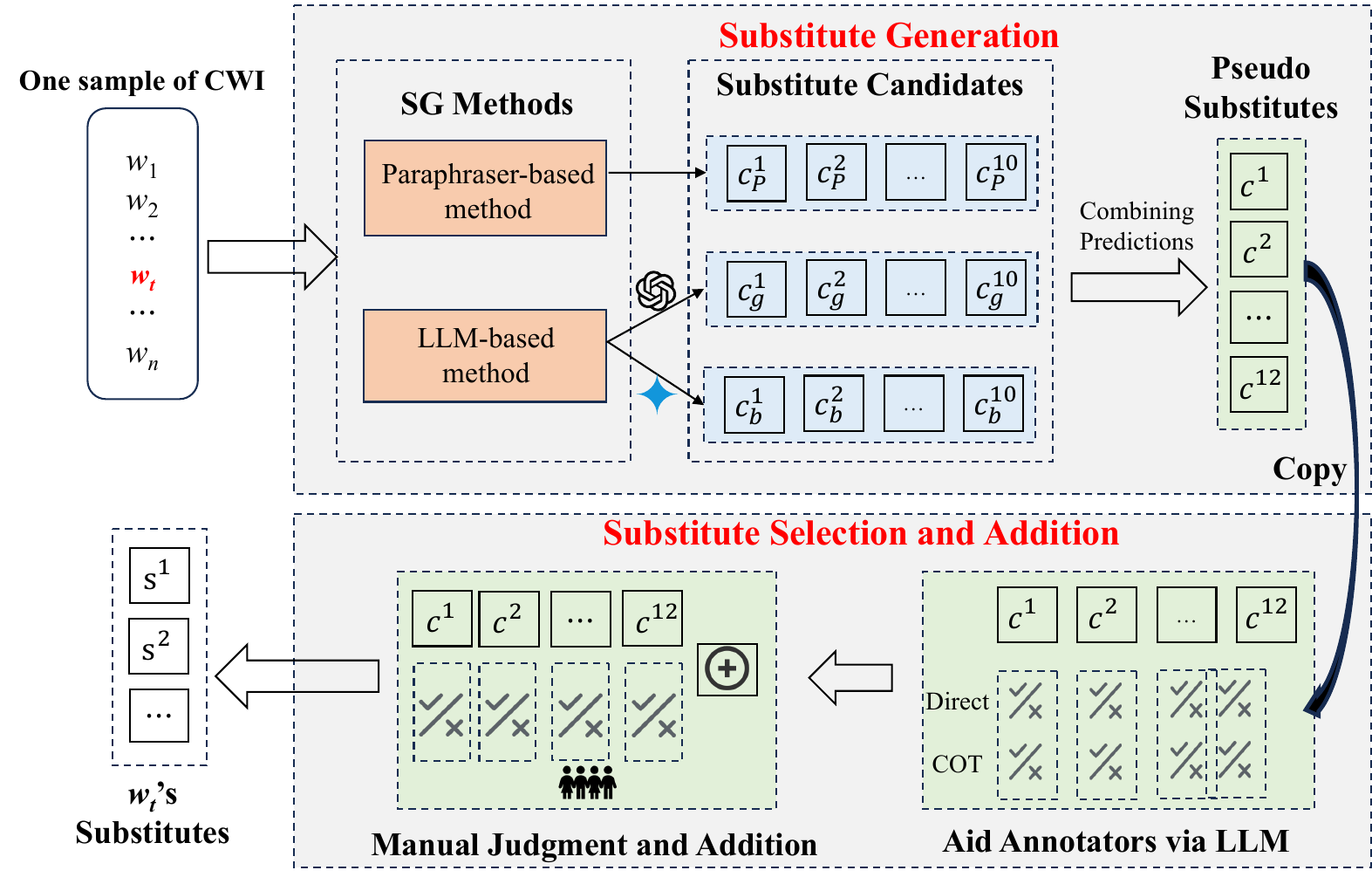}
 \caption{An overview of the methodology of the LS corpus we built. In the substitute generation phase, we combine three different methods via majority vote to generate pseudo substitutes. In the annotation stage, we first use LLMs to perform the first round of annotation using direct prompting and chain-of-thought prompting, and then feedback on the results to the annotator for final judgment and addition.}
 \label{fig:dataset}
\end{figure}

Therefore, we need a dataset that annotates all the complex words that can be simplified in each sentence, along with suitable substitutes for each of those complex words. We propose a novel annotation method to construct an all-in-one LS dataset through human-computer collaboration in Figure \ref{fig:dataset}. Compared to direct manual annotation, the adopted human-machine annotation method is influenced by the following two observations.

(1) Automatic LS methods can offer a broader array of substitutes. By utilizing computational tools such as paraphrasing models or LLMs, an extensive range of viable substitutes can be produced. This variety enhances the dataset by providing numerous substitution choices, capturing diverse semantic relationships and syntactic structures.

(2) Assessing the appropriateness of these substitutes is significantly easier for annotators compared to creating a substitute from memory. Human annotators can focus on selecting the most suitable substitutes from the AI-generated collection, ensuring high-quality and contextually relevant annotations.

\subsection{Data Preparation}

Considering that these CWI datasets from CWI 2018 shared task have already been manually annotated with all complex words in the sentences, we can use these CWI datasets to construct one all-in-one LS dataset, thereby omitting the step of annotating complex words. It includes three distinct
text genres: professionally written news (News), non-professionally written news (WikiNews), and Wikipedia articles (Wikipedia). 

We randomly selected 400 sentences from three topics, removed complex words such as fixed phrases and proper nouns, and obtained sentences with annotated complex words. Why do we need to remove some of the complex words annotated in the dataset? Here, if a word does not have a suitable substitute, it cannot be considered a complex word, such as many nouns. From the perspective of lexical understanding, words that are not recognized by the majority of people are considered complex words. However, from the perspective of lexical simplification, only those complex words for which suitable substitutes can be found are meaningful to simplify, and these are the words considered complex.

\subsection{Substitute Generation} 

Given a sentence of CWI dataset and the corresponding set of complex words, we employ three different LS methods to generate a set of pseudo substitutes for each complex word. Here, we chose one of the best small model-based method LSBert \cite{LiuQLYZH23} and two LLM-based methods (GPT3.5 and Gemini1.0). By utilizing these different methods based on different semantic knowledge, we aim to enhance the overall diversity of the substitutes available for consideration. We use few-shot prompting to guide the models to generate the candidates for the target word from high to low quality.

Suppose the generated substitutes for one complex word $w$ from three LS methods are $\{R_1, R_2, R_3\}$, respectively. Finally, to integrate and re-rank the three results, we assign a combination score for each candidate using Equation~\ref{eq:func2}. We choose the top 12 candidates for $w$ based on the ranking results.

\begin{equation}
    Score(r) = \sum_{i=1}^3f(r)_{R_i}
    \label{eq:func2}
\end{equation}
\begin{equation}
\text{f}(r)_{R_i} = \begin{cases}
    5 - 0.5 \cdot \text{index}(r)_{R_i} & \text{if } r \in R_i\\
    0, & \text{otherwise}
\end{cases}
\label{eq:func3}
\end{equation} 
where $\text{index}(r)_{R_i}$ denotes the index of candidate $r$ in $R_i$, starting from 0. 

\subsection{Manual annotation assisted by LLMs}

After obtaining the substitute candidates, the annotators need to consider two questions: (1) whether the substitute is simpler than the complex word? and (2) whether the meaning of the sentence remains the same after the substitute replaces the complex word? Only if the answer to both questions will be “YES”, the substitute can be chosen as a suitable substitute.

Here, we can directly recruit humans with good English proficiency to judge or add other suitable substitutes. To provide better utilization of the capabilities of LLMs, we continue to offer aid to annotators on the suitability of pseudo substitutes. We choose two LLMs (GPT3.5 and Gemini1.0), combined with few-shot prompting (Direct) and chain-of-thought (COT) prompting strategies, to annotate the pseudo-substitution words. We let LLM evaluate each pseudo substitute to determine if it meets the above two conditions. If it does, it is considered an appropriate substitute; if it does not, it is not considered appropriate.

We developed a dedicated website for data annotation. Each page displays a sentence with a highlighted target word and 12 pseudo substitutes for that word. For each pseudo substitute, four recommended results from two LLMs. Annotators can also add new substitutes not included among the pseudo-substitutes.

Each pseudo substitute has three radio buttons labeled "YES", "NO", and "UNSURE". Annotators need to select "YES" if they consider the substitute a suitable replacement for the target word within the given sentence. They should choose "NO" if the substitute is inappropriate, and select "UNSURE" if they are uncertain. Here, the results recommended by the large model serve as an auxiliary strategy to provide annotators with a reference. For further details on the website and prompts, please see Appendix~\ref{sec:appendix_Dataset}.

\begin{table}[]
    \centering
    \begin{tabular}{l|c|c|ccc}\hline
\multirow{2}{*}{Dataset} & \multirow{2}{*}{NOI} & \multirow{2}{*}{NOC} & \multicolumn{3}{c}{Substitutes}\\
&  & & Min  & Max  & Avg \\\hline 
WikiNews  &100 & 412 & 2 & 13  & 6.1  \\
News  & 150 & 621 & 1 & 12  & 5.0  \\
Wikipedia &150 & 531 & 1 & 12  & 5.4  \\\hline
    \end{tabular}
    \caption{Statistics on the constructed LS dataset. "NOI" is the number of instances, and "NOC" is the number of all complex words in all instances. Columns four to six indicate the minimum, maximum, and average number of substitutes for each complex word}
    \label{tab:dataset_statistics} 
\end{table}

\subsection{Data analysis}

 The statistical information of the constructed LS dataset is shown in Table~\ref{tab:dataset_statistics}. The dataset contains a total of 400 instances, with each instance containing an average of four complex words and each complex word containing an average of five simplified substitutes. 

\begin{table}[]
    \centering
\begin{tabular}{l|c|c}\hline
Dataset  &  3 &  4\\\hline
WikiNews  & 0.804       & 0.930   \\ \hline
News      & 0.763       & 0.904 \\\hline
Wikipedia & 0.766       & 0.900  \\\hline
\end{tabular}
\caption{Consistency test of human annotators with the results from two LLMs. "3" means that LLM only adopted the results where at least 3 out of 4 were the same to comparison, and "4" means that all four results must be the same. }
 \label{tab:consistency} 
\end{table}

\textbf{LLM Usefulness.}  We analyze whether the judgments made by LLMs are useful. We do a consistency test aimed at comparing the annotation results between human annotators and LLMs. For each ranking, we obtained four prediction results from LLMs. Based on majority voting, LLM's different prediction results must be consistent to be adopted. Here, we compared two situations: one where the large model adopted three consistent results and one where it adopted four consistent results.

The results are shown in Table~\ref{tab:consistency}.  We observe that the annotation results show a high degree of overlap. This finding implies that LLM can provide effective information to supplement the annotator's work, thereby reducing their workload and significantly improving the overall efficiency of the labeling process. 

\textbf{High quality.} The objective is to evaluate the accuracy of the substitutions made in the given sentence and complex word pairs. A total of 150 instances were randomly selected, with 50 instances chosen from one of three text genres. A new annotator, proficient in the English language, was assigned the task of assessing the precision of the substitutions within the selected instances. The precision of the substitutions was computed by dividing the number of correct substitutes by the total number of substitutes evaluated. The high precision rate of above 94\% (779/828) indicates the high quality of the substitutions within the dataset. 

\textbf{High coverage.} The substitutes provided by the evaluators are compared against the set of substitutions present in the constructed dataset. The same 150 instances in high quality are selected. Two new human evaluators, proficient in the English language, were asked to independently think of substitutes for each sentence and complex word pair in the selected instances. 

The human annotators provide 342 distinct substitutions and 325 substitutions belonged to the substitutions provided in the dataset. The coverage is equivalent to 95\% (325/342). Additionally, it is worth noting that the number of substitutes 342 is significantly smaller than the 828 substitutes present in the dataset. This observation highlights the impracticality of relying solely on manual annotation.

\begin{figure}
\centering
\begin{boxedminipage}{\columnwidth}
\footnotesize
\#\#\#\#Instruction\#\#\#\#\\
Your task is to simplify the complex words of the input sentence which aims at making it easier for children to read and understand. \\
Don't rewrite the sentence, you MUST ensure that the structure of the simplified sentence matches the structure of the original sentence.\\
$[(examples)]$\\
\#\#\#\#Task\#\#\#\#\\
SENTENCE: $[Input\_sentence]$\\ 
ANSWER: 
\end{boxedminipage}
\caption{Prompt template for one-step LS.}
\label{fig:one-stepLS-4shot}
\end{figure}
\section{Methods}

The LS task can be solved by LLM using a single prompt (Section 4.1). Besides, we propose a novel multi-LLM collaboration method (Section 4.2). 

\subsection{LS via a Single Prompt}

Given a sentence $S$, the task is to directly output a simplified sentence that meets the LS requirements. In-context learning has gained popularity due to its effectiveness and efficiency in leveraging LLMs. This technique aims to enhance the results by leveraging semantically similar examples (few-shot) or utilizing uncertainty and diversity for demonstration refinement and evaluation. The few-shot-based LS prompt is shown in Figure \ref{fig:one-stepLS-4shot}. In our experiments, we select four samples from the dataset as demonstrations.

\begin{figure}
\centering
\begin{boxedminipage}{\columnwidth}
\footnotesize
\#\#\#\#Instruction\#\#\#\#\\
Your task is to simplify the complex words of the input sentence that aims at making it easier for children to read and understand. You MUST follow the given steps while performing this task:\\
STEP 1: Identify all the complex words in the sentence except proper nouns and phrases.\\
STEP 2: Generate the simpler alternatives for every complex word. The alternatives produced cannot change the original meaning of the sentence.\\
STEP 3: Keep non-complex words unchanged, and replace complex words with simpler alternatives.\\
Don't rewrite the sentence, you MUST ensure that the structure of the simplified sentence matches the structure of the original sentence.\\
$[(examples)]$\\
\#\#\#\#Task\#\#\#\#\\
SENTENCE: $[Input\_sentence]$\\ 
ANSWER: 
\end{boxedminipage}
\caption{Prompt template for LLM-based LS method (COT).}
\label{fig:one-stepLS}
\end{figure}
\begin{figure*}[h!]
  \centering
   \includegraphics[width = 0.75\linewidth]{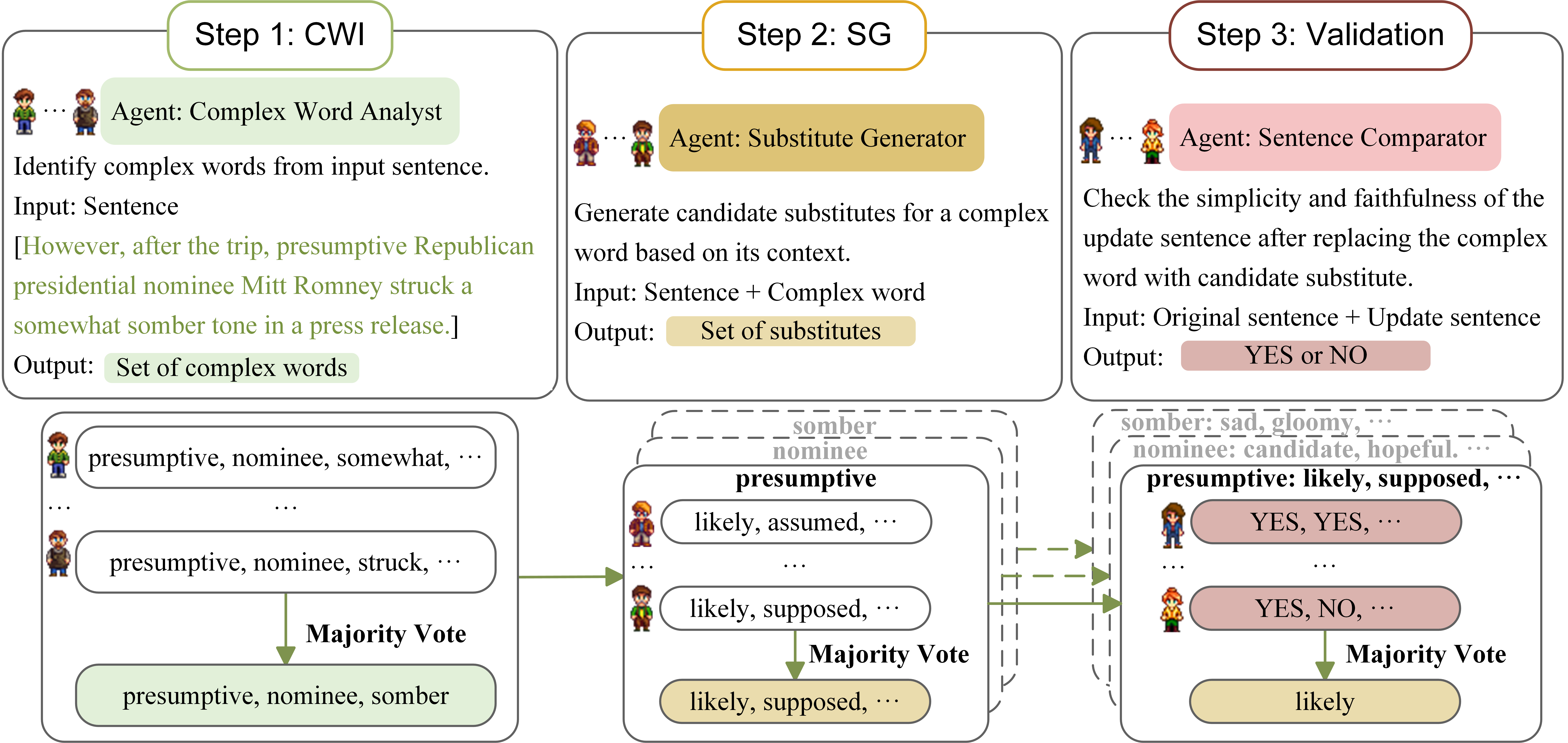}
 \caption{The framework of CoLLS. Each step defines the roles and tasks of LLMs.}
 \label{fig:agent_png}
\end{figure*}
The chain-of-thought (COT) technique \cite{fu2023complexitybased} incorporates reasoning complexity into the demonstration process enabling a more comprehensive understanding of the reasoning process. We also find that guiding the model through COT technique can enhance the quality of text revisions. The COT-based prompt is shown in Figure \ref{fig:one-stepLS}. We need to specify in detail the three steps LS needs to complete and the tasks required for each step. In designing this prompt, we also use relevant examples to enable few-shot learning to improve the performance.

\subsection{Multi-LLM Collaboration}

To address more complex task, research into multi-LLM systems based on LLMs has emerged, demonstrating significant promise \cite{yao2022react,wu2024perhaps,chen2023agentverse}. Therefore, we develop a collaborative framework CoLLS with multiple LLMs, as shown in Figure \ref{fig:agent_png}. 

These LLMs in CoLLS emulate human expert teams to ensure thorough CWI, SG, and Validation. The Validation step here differs from the existing SR step. The SR step involves ranking multiple candidate substitutes, whereas Validation aims to analyze at the sentence level and select the best one. For each LLM, we adopt a majority voting strategy \cite{wang2023selfconsistency,yang2024llmvotinghumanchoices} to improve the overall accuracy and robustness of the predictions. For further details on the prompts of CoLLS, please refer to Appendix~\ref{sec:appendix_b}.

\textbf{CWI.} LLM (Complex Word Analyst) identifies complex words from the input sentence that are relatively difficult to understand. Given $N$ different prompts, we first generate $N$ complex word sets $C=\{C_1,C_2,...,C_N\}$ via LLMs separately and then use a majority voting scheme to determine which complex words occur in at least $m$ of the $N$ predicted outcomes. We define the final complex word set $C_{o}$ as: 

\begin{equation}
C_o = \left\{ w \mid \sum_{i=1}^{N} \mathbf{1}(w \in C_i) \geq m \right\}
\label{equ:cwi}
\end{equation}
where $\mathbf{1}(w \in C_i)$ is the indicator function that is 1 if $w$ is in $C_i$, and 0 otherwise.

\textbf{SG.} LLM (Substitute Generator) generates simplified substitutes for each complex word. We process each complex word in $C_{o}$ individually. For each complex word $w$, we provide a sentence and the complex word in the prompt and let LLM generate candidate substitutes of $w$ based on its context. These candidates should not only be simpler than the complex word but also should not alter the original meaning of the sentence.

We still use the ensemble method via a majority voting scheme to generate simplified substitutes. We refine and formalize this process with the given requirements. Given $N$ different prompts, we generate $N$ candidate substitute sets: $R=\{R_1,R_2,...,R_N\}$.

(1) We use a majority voting scheme to determine which substitutes occur in at least $m$ of the $N$ predicted results as a substitute set $R_o$.

\begin{equation}
R_o = \left\{ r \mid \sum_{i=1}^{N} \mathbf{1}(r \in R_i) \geq m \right\} \\
\end{equation}

(2) For each substitute $r$ in $R_o$, we rank them based on their frequency of occurrence in the combined sets $R$.
\begin{equation}
\text{rank}(r) = \sum_{i=1}^{N} \mathbf{1}(r \in R_i)
\label{equ:rank}
\end{equation}

Through Equation~\ref{equ:rank}, we ensure that the substitution word with the highest votes is most likely to be the substitute for the complex word $w$.

\textbf{Validation.} LLM (Sentence Comparator) finds a more suitable word from $S_o$ from the sentence level. After replacing the complex word of the original sentence with one substitute to form the updated version, we attempt to evaluate the original sentence and the updated sentence. We provide the original sentence and the updated version in the prompt and let LLM check the simplicity, faithfulness, and fluency of the updated version.

When inputting the original sentence and the updated sentence as prompt, we let LLMs make a rationality judgment, and output "Yes" or "No". Here, "Yes" means it meets the requirements, and "No" means it does not. The majority voting strategy is still adopted. Suppose $N$ LLMs are set up, and each of the substitutes from the previous step is validated by $N$ LLMs in sequence, we keep the ones agreed upon by more than $m$ models. Finally, we chose the substitute that received the most "Yes" votes. In case of a tie, we select the top-ranked substitute. If no substitute meets the requirement of $m$ "Yes" votes, the complex word is abandoned.
\begin{table*}[h!]
    \centering
    \begin{tabular}{c|c|ccccc}\hline
  Dataset&Method&NumCW&CorrectCW&CorrectSimp&F1 &F1-20 \\
          \hline
\multirow{9}{*}{WikiNews}
&LSBert                            & \textbf{400}        & \textbf{320}  & 173  & 0.397    & 0.139   \\
\cdashline{2-7}
&GPT3.5                    & 367        & 229  & 203  & 0.524    & 0.207   \\
&GPT3.5(COT)              & 318        & 221  & 204  & 0.541    & 0.210   \\
\cdashline{2-7}
&Llama3       & 347        & 229  & 189  & 0.484    & 0.190   \\
&Llama3(COT) & 317        & 217  & 186  & 0.486    & 0.190  \\
\cdashline{2-7}
&CoLLS(Llama3)           & 260       & 219  & 187  &0.576    & 0.242  \\
&CoLLS(GPT3.5)           & 335       & 258  & \textbf{223}  &\textbf{0.605}    & \textbf{0.247}
\\
\hline
\multirow{9}{*}{News}
&LSBert                            & \textbf{612}   & \textbf{486}  & 268  & 0.434    & 0.180   \\
\cdashline{2-7}
&GPT3.5                    & 555   & 373  & 295  & 0.489    & 0.199   \\
&GPT3.5(COT)              & 582   & 381  & 326  & 0.524    & 0.225   \\
\cdashline{2-7}
&Llama3       & 630   & 407  & 307  & 0.480    & 0.203   \\
&Llama3(COT) & 597   & 395  & 304  & 0.487    & 0.205  \\
\cdashline{2-7}
&CoLLS(Llama3)          & 458    & 370  & 305  &0.583    & 0.278\\
&CoLLS(GPT3.5)           & 578       &439  & \textbf{351}  &\textbf{0.611}    & \textbf{0.280}
\\
\hline
\multirow{9}{*}{Wikipedia}
&LSBert                            & \textbf{711}        & \textbf{457}  & 240  & 0.418    & 0.141   \\
\cdashline{2-7}
&GPT3.5                    & 439        & 282  & 238  & 0.482    & 0.159   \\
&GPT3.5(COT)              & 439        & 288  & 252  & 0.523    & 0.169   \\
\cdashline{2-7}
&Llama3       & 557        & 315  & 238  & 0.444    & 0.159   \\
&Llama3(COT) & 530        & 304  & 238  & 0.450    & 0.160  \\
\cdashline{2-7}
&CoLLS(Llama3)           & 468       & 330  & 266  &0.543    & \textbf{0.210}\\
&CoLLS(GPT3.5)           & 591      & 385  & \textbf{306}  &\textbf{0.569}    & 0.204
\\
\hline
    \end{tabular}
\caption{Results of LS methods on three datasets. NumCW is the number of complex words identified by LS, CorrectCW is the number of complex words correctly identified by LS, CorrectSimp is the number of complex words correctly simplified by LS.}
    \label{tab:result1} 
\end{table*}

\section{Experiments}

\subsection{Experiment Setup}
\textbf{Metrics.} Considering that even when annotated as complex words, the complexity of the words themselves varies; some words may be difficult for many people, while others are only hard to understand for a few. Therefore, when evaluating the performance of the simplification methods, we considered two scenarios.

Given a sentence with $P$ complex words that need to be simplified, the model identifies $Q$ complex words, and the number of correctly simplified words is $q$. 

(1) We do not consider the difficulty differences of complex words—if one person deems it complex, it is considered complex. 

Precision ($\frac{q}{Q}$) refers to the proportion of generated simplified words that are correct. Recall ($\frac{q}{P}$) refers to the proportion of correctly simplified words to all words that need to be simplified. F1 refers to the harmonic mean between Precision and Recall.

(2) We consider the difficulty differences of complex words. When evaluating the effectiveness of simplification methods, if both methods successfully simplify a word, the method that simplifies a word marked as complex by 10 people is considered more effective than one that simplifies a word marked as complex by only 2 people.

Precision is changed as follows,
\begin{equation}
   Precision = \sum_{i=1}^q\frac{H_i}{Q\times20}
    \label{eq:precision_2}
\end{equation}
where $H_i$ indicates how many out of 20 annotators consider the word to be complex When a complex word is correctly simplified.

Recall is changed as follows,
\begin{equation}
   Recall = \sum_{i=1}^q\frac{H_i}{P\times20}
    \label{eq:recall_1}
\end{equation}

F1 is calculated based on the modified Precision and Recall, denoted as \textbf{F1-20}. Compared to F1 without considering the difficulty difference, F1-20 is a more reasonable metric because it takes into account the difficulty differences of complex words.

\textbf{Baselines.} Although there are many LS methods, they treat CWI and SG as separate tasks, so they cannot directly output simplified sentences \cite{LiuQLYZH23}.

\textbf{LSBert.} It is one of the best LS methods, which consists of a network for complex word identification by fine-tuning BERT and a network for substitute generation based on BERT \cite{qiang2021lsbert}. LSBert adopts a supervised method on CWI task.

\textbf{LLMs(GPT3.5 and Llama3).} To evaluate the performance of different LLMs on LS tasks, we choose one open-source model Llama3(Meta-Llama-3-70B-Instruct), and one closed-source model GPT3.5 (GPT-3.5-turbo-1106). GPT3.5 is accessed through their respective provided API services, and Llama3 is accessed through the Ollama package. We adopt two strategies introduced in Section 3: one based on few-shot learning and one based on COT reasoning. We choose four samples for the demonstrations.

\textbf{CoLLS.} We use two LLMs mentioned above to construct CoLLS separately. Specifically, all three steps of CoLLS used $N$ equal to 3 and $m$ equal to 2. To ensure diversity, the number of demonstrations in few-shot settings for the three LLMs at each step are 2, 4, and 6, respectively.

\textbf{AgentLS.} We use two LLMs mentioned above to construct AgentLS separately. Specifically, all three steps of AgentLS used $N$ equal to 3 and $m$ equal to 2. To ensure diversity, the number of demonstrations in few-shot settings for the three LLMs at each step are 2, 4, and 6, respectively.

\subsection{Results}

Table~\ref{tab:result1} shows the results of LS methods. LSBert employs a supervised CWI method, which allows it to identify complex words better. However, we also found that LSBert does not perform well in correctly simplifying them. In contrast, the LLM-based methods can achieve the task without the need for fine-tuning.

Compared to LSBert based on small pretrained modeling, LS methods based on LLMs perform better in LS task, even when using only a single prompt. This redefines the LS task, making it possible to complete the task without dividing it into multiple sub-steps. The use of COT-based LLM method brings a slight performance improvement. The closed-source model GPT3.5 slightly outperforms the open-source model Llama 3.

The multi-LLM collaboration method CoLLS significantly outperforms other LLM-based methods, whether based on Llama 3.0 or GPT3.5. This also confirms that the multi-LLM collaboration approach is more effective in completing LS task. The main reasons can be summarized in the following two points.

(1) Enhanced Decision-Making Through Collaboration: CoLLS leverages the strengths of multiple LLMs working together, which allows for more thorough and balanced decision-making. By using multiple models to validate each potential simplification, CoLLS reduces the likelihood of errors that might arise from relying on a single model. This collaborative approach ensures that only the most accurate and contextually appropriate simplifications are selected, leading to better overall performance.

(2) Refinement Through Majority Voting: CoLLS employs a majority voting, where only simplifications that are agreed upon by a majority of the models are considered. This consensus-based approach helps filter out less optimal or contextually inappropriate simplifications, ensuring that the final output is of higher quality. The use of majority voting also adds a layer of robustness, making CoLLS more reliable in producing consistent and accurate simplifications.

\subsection{Ablation Study}

In each step of CoLLS, we adopted a majority voting strategy, with the default setting being 3 out of 2, i.e., $N$ equals 3 and $M$ equals 2. Here, we explore the impact of $N$ on the performance of CoLLS. Specifically, in each of the three steps, we change one step at a time while keeping the other two fixed with $N$ equal to 3 and $M$ equal to 2. In the varied step, the value of $N$ ranges from 1 to 6, and the value of $M$ is set to half of $N$, meaning that a decision is accepted as long as half or more agree, where 1 means that no majority voting is used. Additionally, when conducting ablation experiments on the Validation step, we also included the case where N equals 0. This situation indicates that no Validation is performed, and the substitute ranked highest in the SG is directly used as the substitute.

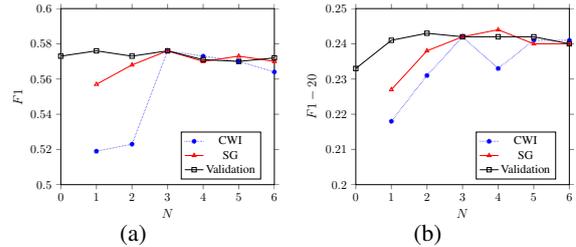
\begin{figure}[]
\subfigure[]{
\pgfplotsset{compat=1.3}
\begin{tikzpicture}[scale=0.41,baseline] 
\begin{axis}[
    xlabel=$N$,
    xmin=0,
    xmax=6,
    ylabel=$F1$,
    ymin=0.5,
    ymax=0.6,
    label style={font=\large},
    tick align=outside, 
    legend style={at={(0.75,0.3)},anchor=north,font=\normalsize} 
    ]

\addplot[sharp plot,mark=*,blue,densely dotted] plot coordinates { 
    (1,0.519)
    (2,0.523)
    (3,0.576)
    (4,0.573)
    (5,0.570)
    (6,0.564)
};
\addlegendentry{CWI}
\addplot[sharp plot,mark=triangle,red] plot coordinates { 
    (1,0.557)
    (2,0.568)
    (3,0.576)
    (4,0.570)
    (5,0.573)
    (6,0.570)
};
\addlegendentry{SG}
\addplot[sharp plot,mark=square,black] plot coordinates {
    (0,0.573)
    (1,0.576)
    (2,0.573)
    (3,0.576)
    (4,0.571)
    (5,0.570)
    (6,0.572)
};
\addlegendentry{Validation}
\end{axis}
\end{tikzpicture}}
\subfigure[]{
\pgfplotsset{compat=1.3}
\begin{tikzpicture}[scale=0.41,baseline] 
\begin{axis}[
    xlabel=$N$, 
    xmin=0,
    xmax=6,
    ylabel=$F1-20$, 
    ymin=0.2,
    ymax=0.25,
    label style={font=\large},
    tick align=outside, 
    legend style={at={(0.75,0.3)},anchor=north,{font=\normalsize}} 
    ]
\addplot[sharp plot,mark=*,blue,densely dotted] plot coordinates { 
    (1,0.218)
    (2,0.231)
    (3,0.242)
    (4,0.233)
    (5,0.241)
    (6,0.241)
};
\addlegendentry{CWI}
\addplot[sharp plot,mark=triangle,red] plot coordinates {
    (1,0.227)
    (2,0.238)
    (3,0.242)
    (4,0.244)
    (5,0.240)
    (6,0.240)
};
\addlegendentry{SG}
\addplot[sharp plot,mark=square,black] plot coordinates {
    (0,0.233)
    (1,0.241)
    (2,0.243)
    (3,0.242)
    (4,0.242)
    (5,0.242)
    (6,0.240)
};
\addlegendentry{Validation}
\end{axis}
\end{tikzpicture}}

\caption{Effect of varying the number of LLMs ($N$) for CoLLS(Llama3) on WikiNews. (a) the results on metric F1, and (b) the results on metric F1-20.} 
\label{fig:varyN}
\end{figure}

We selected the WikiNews dataset to analyze the performance of CoLLS(Llama3), and the results are shown in Figure \ref{fig:varyN}. It can be observed that when $N$ is set to 1, CoLLS performs the worst. When N is set to 3 or higher, the results stabilize. This indicates that the majority voting strategy is effective. Regarding the Validation step (when N equals 0), the F1 value does not change significantly, but the F1-20 value is somewhat lower. This suggests that the Validation step can enhance performance.
\subsection{Conclusions}

We found Large Language Models (LLMs) can directly generate simplified sentences in a single step, without needing to simplify through a pipeline like existing methods do.  Therefore, we focus on a new evaluation paradigm for lexical simplification task. In this paer, we introduced a novel annotation method that combines human and machine efforts to create a all-in-one LS dataset. We also explored multiple LLM collaboration within LLMs, which outperformed single-prompt methods, advancing LS research and expanding the potential of multi-LLM systems in NLP. In the future, we have only constructed the English dataset here, we will construct the all-in-one LS dataset for other languages.

\section*{Limitations}
While the collaborative approach we proposed successfully constructed an all-in-one style LS dataset, we must acknowledge certain limitations in order to provide a balanced perspective.

The coverage of the dataset is limited to two genres (Wiki and News), which may affect its applicability in certain contexts. Researchers should exercise caution when generalizing findings beyond the scope of the dataset. In order to efficiently complete the experiments within the constraints of limited computational resources and time, we constructed a dataset of only 400 instances. This may not fully capture the diversity of linguistic phenomena. Future research will expand the size and scope of the dataset to include a broader range of text types and linguistic phenomena, thereby ensuring the model's effectiveness across various application scenarios.

\bibliography{ref}

\begin{thebibliography}{29}
\providecommand{\natexlab}[1]{#1}

\bibitem[{Aumiller and Gertz(2022)}]{aumiller-gertz-2022-unihd}
Dennis Aumiller and Michael Gertz. 2022.
\newblock {U}ni{HD} at {TSAR}-2022 shared task: Is compute all we need for lexical simplification?
\newblock In \emph{Proceedings of the Workshop on Text Simplification, Accessibility, and Readability (TSAR-2022)}, pages 251--258.

\bibitem[{Bott et~al.(2012)Bott, Rello, Drndarevic, and Saggion}]{bott2012Can}
S~Bott, L~Rello, B~Drndarevic, and H.~Saggion. 2012.
\newblock Can spanish be simpler? lexsis: Lexical simplification for spanish.
\newblock In \emph{Proceedings of COLING}, pages 357--374.

\bibitem[{Chen et~al.(2023)Chen, Su, Zuo, Yang, Yuan, Qian, Chan, Qin, Lu, Xie et~al.}]{chen2023agentverse}
Weize Chen, Yusheng Su, Jingwei Zuo, Cheng Yang, Chenfei Yuan, Chen Qian, Chi-Min Chan, Yujia Qin, Yaxi Lu, Ruobing Xie, et~al. 2023.
\newblock Agentverse: Facilitating multi-agent collaboration and exploring emergent behaviors in agents.
\newblock \emph{arXiv preprint arXiv:2308.10848}.

\bibitem[{Fu et~al.(2023)Fu, Peng, Sabharwal, Clark, and Khot}]{fu2023complexitybased}
Yao Fu, Hao Peng, Ashish Sabharwal, Peter Clark, and Tushar Khot. 2023.
\newblock \href {https://arxiv.org/abs/2210.00720} {Complexity-based prompting for multi-step reasoning}.
\newblock \emph{Preprint}, arXiv:2210.00720.

\bibitem[{Glava{\v{s}} and {\v{S}}tajner(2015)}]{glavavs2015simplifying}
Goran Glava{\v{s}} and Sanja {\v{S}}tajner. 2015.
\newblock Simplifying lexical simplification: do we need simplified corpora?
\newblock In \emph{ACL}, pages 63--68.

\bibitem[{Gooding and Kochmar(2018)}]{gooding2018camb}
Sian Gooding and Ekaterina Kochmar. 2018.
\newblock Complex word identification with ensemble-based voting.
\newblock In \emph{In Proceedings of the Thirteenth Workshop on Innovative Use of NLP for Building Educational Applications}, pages 184--194.

\bibitem[{Gooding and Kochmar(2019)}]{gooding2019complex}
Sian Gooding and Ekaterina Kochmar. 2019.
\newblock Complex word identification as a sequence labelling task.
\newblock In \emph{Proceedings of the 57th Annual Meeting of the Association for Computational Linguistics}, pages 1148--1153.

\bibitem[{Liu et~al.(2023)Liu, Qiang, Li, Yuan, Zhu, and Hua}]{LiuQLYZH23}
Kang Liu, Jipeng Qiang, Yun Li, Yunhao Yuan, Yi~Zhu, and Kaixun Hua. 2023.
\newblock Multilingual lexical simplification via paraphrase generation.
\newblock In \emph{26th European Conference on Artificial Intelligence,}, volume 372, pages 1529--1535.

\bibitem[{Maddela and Xu(2018)}]{maddela-xu-2018-word}
Mounica Maddela and Wei Xu. 2018.
\newblock A word-complexity lexicon and a neural readability ranking model for lexical simplification.
\newblock In \emph{EMNLP}, pages 3749--3760.

\bibitem[{North et~al.(2024)North, Ranasinghe, Shardlow, and Zampieri}]{north2024multils}
Kai North, Tharindu Ranasinghe, Matthew Shardlow, and Marcos Zampieri. 2024.
\newblock Multils: A multi-task lexical simplification framework.
\newblock \emph{arXiv preprint arXiv:2402.14972}.

\bibitem[{North et~al.(2023)North, Zampieri, and Shardlow}]{north2023lexical}
Kai North, Marcos Zampieri, and Matthew Shardlow. 2023.
\newblock Lexical complexity prediction: An overview.
\newblock \emph{ACM Computing Surveys}, 55(9):1--42.

\bibitem[{Paetzold and Specia(2017{\natexlab{a}})}]{paetzold2017lexical}
Gustavo Paetzold and Lucia Specia. 2017{\natexlab{a}}.
\newblock Lexical simplification with neural ranking.
\newblock In \emph{ACL: Volume 2, Short Papers}, pages 34--40.

\bibitem[{Paetzold and Specia(2016)}]{paetzold2016unsupervised}
Gustavo~H Paetzold and Lucia Specia. 2016.
\newblock Unsupervised lexical simplification for non-native speakers.
\newblock In \emph{AAAI}, pages 3761--3767.

\bibitem[{Paetzold and Specia(2017{\natexlab{b}})}]{paetzold2017survey}
Gustavo~H Paetzold and Lucia Specia. 2017{\natexlab{b}}.
\newblock A survey on lexical simplification.
\newblock In \emph{Journal of Artificial Intelligence Research}, volume~60, pages 549--593.

\bibitem[{Pavlick and Callison-Burch(2016)}]{pavlick2016simple}
Ellie Pavlick and Chris Callison-Burch. 2016.
\newblock Simple ppdb: A paraphrase database for simplification.
\newblock In \emph{ACL: Volume 2, Short Papers}, pages 143--148.

\bibitem[{Qiang et~al.(2021{\natexlab{a}})Qiang, Li, Zhu, Yuan, Shi, and Wu}]{qiang2021lsbert}
Jipeng Qiang, Yun Li, Yi~Zhu, Yunhao Yuan, Yang Shi, and Xindong Wu. 2021{\natexlab{a}}.
\newblock Lsbert: Lexical simplification based on bert.
\newblock \emph{IEEE/ACM Transactions on Audio, Speech, and Language Processing}, 29:3064--3076.

\bibitem[{Qiang et~al.(2020)Qiang, Li, Zhu, Yuan, and Wu}]{qiang2020AAAI}
Jipeng Qiang, Yun Li, Yi~Zhu, Yunhao Yuan, and Xindong Wu. 2020.
\newblock Lexical simplification with pretrained encoders.
\newblock \emph{Thirty-Fourth AAAI Conference on Artificial Intelligence}, page 8649–8656.

\bibitem[{Qiang et~al.(2021{\natexlab{b}})Qiang, Lv, Li, Yuan, and Wu}]{qiang2021chinese}
Jipeng Qiang, Xinyu Lv, Yun Li, Yunhao Yuan, and Xindong Wu. 2021{\natexlab{b}}.
\newblock Chinese lexical simplification.
\newblock \emph{IEEE Transactions on Audio, Speech and Language Processing.}, 29:1819--1828.

\bibitem[{Saggion(2017)}]{saggion2017automatic}
Horacio Saggion. 2017.
\newblock Automatic text simplification.
\newblock \emph{Synthesis Lectures on Human Language Technologies}, 10(1):1--137.

\bibitem[{Seneviratne and Suominen(2024)}]{seneviratne-suominen-2024-anu}
Sandaru Seneviratne and Hanna Suominen. 2024.
\newblock {ANU} at {MLSP}-2024: Prompt-based lexical simplification for {E}nglish and {S}inhala.
\newblock In \emph{Proceedings of the 19th Workshop on Innovative Use of NLP for Building Educational Applications (BEA 2024)}, pages 599--604.

\bibitem[{Shardlow et~al.(2021)Shardlow, Evans, Paetzold, and Zampieri}]{shardlow-etal-2021-semeval}
Matthew Shardlow, Richard Evans, Gustavo~Henrique Paetzold, and Marcos Zampieri. 2021.
\newblock {S}em{E}val-2021 task 1: Lexical complexity prediction.
\newblock In \emph{Proceedings of the 15th International Workshop on Semantic Evaluation (SemEval-2021)}, pages 1--16.

\bibitem[{Sheang et~al.(2022)Sheang, Ferr{\'e}s, and Saggion}]{sheang-etal-2022-controllable}
Kim~Cheng Sheang, Daniel Ferr{\'e}s, and Horacio Saggion. 2022.
\newblock Controllable lexical simplification for {E}nglish.
\newblock In \emph{Proceedings of the Workshop on Text Simplification, Accessibility, and Readability (TSAR-2022)}, pages 199--206.

\bibitem[{Sheang and Saggion(2023)}]{sheang2023multilingual}
Kim~Cheng Sheang and Horacio Saggion. 2023.
\newblock Multilingual controllable transformer-based lexical simplification.
\newblock \emph{arXiv preprint arXiv:2307.02120}.

\bibitem[{Wang et~al.(2023)Wang, Wei, Schuurmans, Le, Chi, Narang, Chowdhery, and Zhou}]{wang2023selfconsistency}
Xuezhi Wang, Jason Wei, Dale Schuurmans, Quoc~V Le, Ed~H. Chi, Sharan Narang, Aakanksha Chowdhery, and Denny Zhou. 2023.
\newblock \href {https://openreview.net/forum?id=1PL1NIMMrw} {Self-consistency improves chain of thought reasoning in language models}.
\newblock In \emph{The Eleventh International Conference on Learning Representations}.

\bibitem[{Wu et~al.(2024)Wu, Yuan, Haffari, and Wang}]{wu2024perhaps}
Minghao Wu, Yulin Yuan, Gholamreza Haffari, and Longyue Wang. 2024.
\newblock (perhaps) beyond human translation: Harnessing multi-agent collaboration for translating ultra-long literary texts.
\newblock \emph{arXiv preprint arXiv:2405.11804}.

\bibitem[{Wubben et~al.(2012)Wubben, Bosch, and Krahmer}]{wubben2012}
Sander Wubben, Antal Van~Den Bosch, and Emiel Krahmer. 2012.
\newblock Sentence simplification by monolingual machine translation.
\newblock In \emph{EMNLP}, pages 1015--1024.

\bibitem[{Yang et~al.(2024)Yang, Dailisan, Korecki, Hausladen, and Helbing}]{yang2024llmvotinghumanchoices}
Joshua~C. Yang, Damian Dailisan, Marcin Korecki, Carina~I. Hausladen, and Dirk Helbing. 2024.
\newblock \href {https://arxiv.org/abs/2402.01766} {Llm voting: Human choices and ai collective decision making}.
\newblock \emph{Preprint}, arXiv:2402.01766.

\bibitem[{Yao et~al.(2022)Yao, Zhao, Yu, Du, Shafran, Narasimhan, and Cao}]{yao2022react}
Shunyu Yao, Jeffrey Zhao, Dian Yu, Nan Du, Izhak Shafran, Karthik Narasimhan, and Yuan Cao. 2022.
\newblock React: Synergizing reasoning and acting in language models.
\newblock \emph{arXiv preprint arXiv:2210.03629}.

\bibitem[{Yimam et~al.(2018)Yimam, Biemann, Malmasi, Paetzold, Specia, {\v{S}}tajner, Tack, and Zampieri}]{yimam2018report}
Seid~Muhie Yimam, Chris Biemann, Shervin Malmasi, Gustavo~H Paetzold, Lucia Specia, Sanja {\v{S}}tajner, Ana{\"\i}s Tack, and Marcos Zampieri. 2018.
\newblock A report on the complex word identification shared task 2018.
\newblock pages 66--78.

\end{thebibliography}

\appendix

\section{Dataset construction}
\label{sec:appendix_Dataset}
\subsection{Annotation Website}
To enhance the efficiency of the annotation process, we developed an annotation platform based on Vue+FastAPI. On this platform, annotators will receive a sentence containing a target word, which is marked by ``\textless\textless\textgreater\textgreater", each sentence is accompanied by a set of pseudo substitutes and preliminary annotations from LLMs. 

For each pseudo substitute, annotators are required to evaluate the simplicity and faithfulness of the word. If a pseudo substitute is simpler than the target word and the sentence meaning remains unchanged after replacing the target word, the annotator should check "YES" in the OPTION column. If both conditions are not met, they should check "NO", and if uncertain, they should check "UNSURE". Additionally, if annotators believe there are more appropriate simplified words not included in the 12 pseudo substitutes, we encourage them to add these words in the text box to enrich and improve our dataset.
\begin{figure}[t!]
  \centering
   \includegraphics[width=78mm]{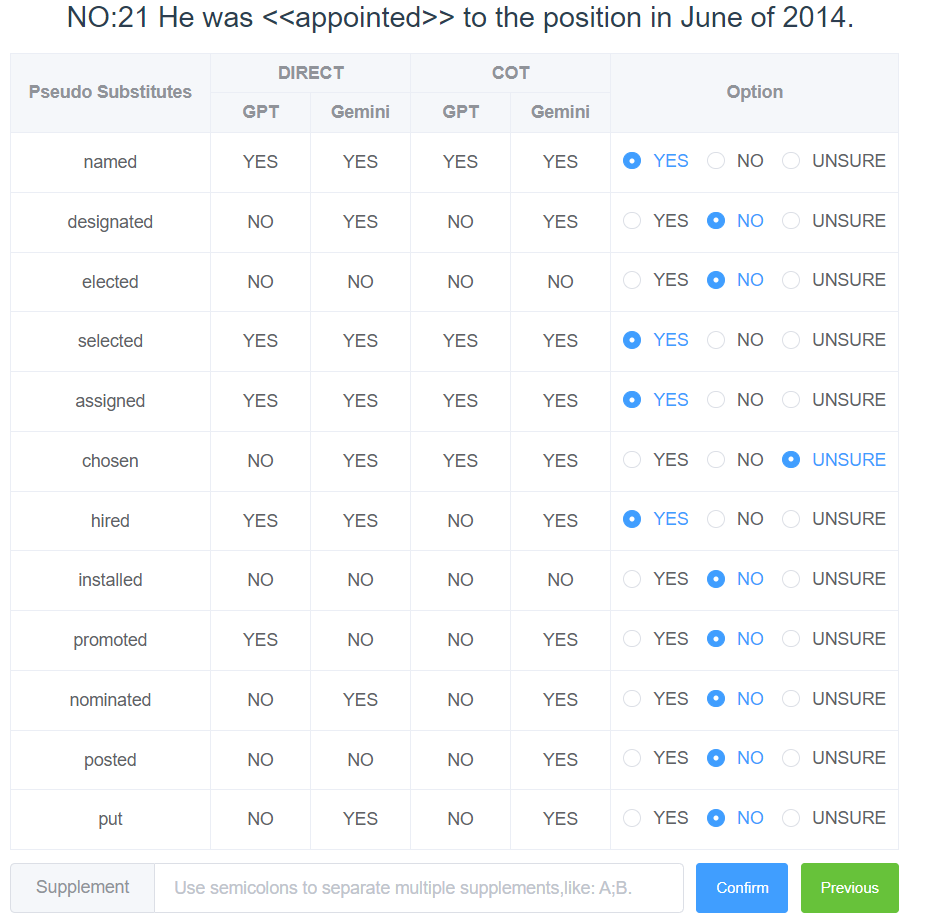}
 \caption{Screenshot of an annotation example on the annotation Website. "Direct" and "COT" are the recommended results from LLMs.}
 \label{website}
\end{figure}

The final dataset comprises 400 sentences, with an average of four complex words per sentence and an average of five simplified words for each complex word.

\subsection{Prompts of constructing the dataset }

In the process of constructing the LS dataset, we used three prompts: the prompt for generating candidate words is shown in Figure \ref{fig:agent_simp}, few-shot prompting (Direct) in Figure \ref{fig:LLM_annotation} and chain-of-thought (COT) prompting in Figure \ref{fig:LLM_annotation_COT} for offering aid annotators on suitability of pseudo substitutes.

\begin{figure}[H]
\centering
\begin{boxedminipage}{\columnwidth}
\footnotesize
\#\#\#\#Instruction\#\#\#\#\\
You are given a sentence in which a target word is surrounded by ``\textless\textless\textgreater\textgreater". You need to provide 10 words that are simpler than the target word and do not change the meaning of the sentence when you replace the target word.\\
Returns one line of results, separated by ";" between the simplified words, do not return any other text!\\
$[(examples)]$\\
\#\#\#\#Task\#\#\#\#\\
SENTENCE: [sentence]\\
TARGET: [target]\\
ANSWER: 
\end{boxedminipage}
\caption{Prompt template of Substitute Generator in Dataset and CoLLS.}
\label{fig:agent_simp}
\end{figure}

\begin{figure}[H]
\centering
\begin{boxedminipage}{\columnwidth}
\footnotesize
\#\#\#\#Instruction\#\#\#\#\\
You are given a sentence in which a target word is surrounded by ``\textless\textless\textgreater\textgreater", and your task is to first determine whether the meaning of a sentence remains the same after replacing the target word with the given alternative, and then to determine whether the given alternative is easier to understand for children than the target word. If both of the above conditions are met, return \#\#YES\#\#. Otherwise, return \#\#NO\#\#.\\
\#\#\#\#Example 1\#\#\#\#\\
SENTENCE: The text is an \textless\textless indication\textgreater\textgreater that it was premeditated, Goodyear said.\\
ALTERNATIVE: sign\\
ANSWER: \#\#YES\#\#\\
...\\
\#\#\#\#Task\#\#\#\#\\
SENTENCE: [sentence]\\
ALTERNATIVE: [alternative]\\
ANSWER:
\end{boxedminipage}
\caption{Prompt template (Direct) for Manual annotation assisted by LLMs.}
\label{fig:LLM_annotation}
\end{figure}

\begin{figure}
\centering
\begin{boxedminipage}{\columnwidth}
\footnotesize
\#\#\#\#Instruction\#\#\#\#\\
You are given a sentence in which a target word is surrounded by ``\textless\textless\textgreater\textgreater", and your task is to first determine whether the meaning of a sentence remains the same after replacing the target word with the given alternative, and then to determine whether the given alternative is easier to understand for children than the target word. If both of the above conditions are met, return \#\#YES\#\#. Otherwise, return \#\#NO\#\#. \\
\#\#\#\#Example 1\#\#\#\#\\
SENTENCE: The text is an \textless\textless indication\textgreater\textgreater that it was premeditated, Goodyear said.\\
ALTERNATIVE: sign\\
ANSWER: Meaning analysis: Both "indication" and "sign" in this context convey a similar idea of something that suggests or reveals information. So the condition "the meaning remains the same after replacing the target word with the given alternative" is satisfied.
Understanding for Children: "sign" is generally simpler and more commonly used in everyday language, which makes it easier for children to grasp compared to "indication". So the condition "the given alternative is easier to understand for children than the target word" is satisfied.
After analysis, since both conditions are satisfied, the answer is \#\#YES\#\#.\\
...\\
\#\#\#\#Task\#\#\#\#\\
SENTENCE: [sentence]\\
ALTERNATIVE: [alternative]\\
ANSWER:
\end{boxedminipage}
\caption{Prompt template (Direct) for Manual annotation assisted by LLMs.}
\label{fig:LLM_annotation_COT}
\end{figure}

\section{The prompt templates of CoLLS}
\label{sec:appendix_b}
The prompt templates of CWI, SG, Validation are shown in Figure \ref{fig:agent_find_complex_word}, \ref{fig:agent_simp}, \ref{fig:agent_val_sentence_level}, respectively. 

\begin{figure}[H]
\centering
\begin{boxedminipage}{\columnwidth}
\footnotesize
\#\#\#\#Instruction\#\#\#\#\\
Identify ALL words in the sentence that are not easy for the child to understand and read; words cannot be proper nouns or phrases made up of more than one word, and complex words are separated by a ";".Just return the complex words with NO other texts.\\
\#\#\#\#Example 1\#\#\#\#\\
SENTENCE: According to police, the two fatalities do not have any obvious connection to the host, but did know each other.\\
ANSWER: fatalities;obvious;connection\\
...\\
\#\#\#\#Task\#\#\#\#\\
SENTENCE: [sentence]\\
ANSWER: 
\end{boxedminipage}
\caption{Prompt template of CoLLS (Complex word analyst).}
\label{fig:agent_find_complex_word}
\end{figure}

\begin{figure}[H]
\centering
\begin{boxedminipage}{\columnwidth}
\footnotesize
\#\#\#\#Instruction\#\#\#\#\\
You are an editor with a solid writing foundation and extensive reviewing experience, given an original SENTENCE 1, and its simplified SENTENCE 2, compare whether SENTENCE 2 has the same meaning as SENTENCE 1; whether SENTENCE 2 is simpler than SENTENCE 1, and whether SENTENCE 2 has the same or better fluency as SENTENCE 1. If all of the above are satisfied, return \#\#YES\#\#, otherwise return \#\#NO\#\#. 
Do not return any text other than the answer.\\
\#\#\#\#Task\#\#\#\#\\
SENTENCE 1: [sentence1]\\
SENTENCE 2: [sentence2]\\
ANSWER:
\end{boxedminipage}
\caption{Prompt template of CoLLS (Sentence comparator).}
\label{fig:agent_val_sentence_level}
\end{figure}

\section{Case study}
\label{sec:case}
Here, we randomly choose 5 instances from WikiNews for analysis in Table \ref{tab:instances}.

\begin{table*}[]
\begin{tabular}{l|l}
\hline
\textbf{Inst.1} & He was \textbf{appointed} to the \textbf{position} in June of 2014.\\ \hline
LSBert          & He was \textcolor{blue}{elected} to the \textcolor{red}{position} in June of 2014.\\
GPT3.5          & He \textcolor{red}{got} the \textcolor{red}{job} in June of 2014.\\
GPT3.5(COT)     & He was \textcolor{red}{assigned} to the \textcolor{blue}{position} in June of 2014.\\
Llama3          & He was \textcolor{blue}{appointed} to the \textcolor{red}{job} in June of 2014.\\
Llama3(COT)     & He was \textcolor{blue}{appointed} to the \textcolor{red}{job} in June of 2014.\\
CoLLS(Llama3) & He was \textcolor{red}{named} to the \textcolor{red}{job} in June of 2014.\\
CoLLS(GPT3.5) & He was \textcolor{red}{selected} to the \textcolor{red}{job} in June of 2014.\\ \hline
\textbf{Inst.2}          & "\textbf{Eventually} after many \textbf{incidents} of his \textbf{anger} coming to the fore, we \textbf{dismissed} him.\\\hline
LSBert          & "\textcolor{blue}{then} after many \textcolor{red}{cases} of his \textcolor{red}{rage} coming to the \textcolor{blue}{front}, we \textcolor{blue}{dismissed} him.\\
GPT3.5          & "\textcolor{red}{Finally} after many \textcolor{red}{events} of his \textcolor{red}{rage} coming to the \textcolor{blue}{front}, we \textcolor{red}{fired} him.\\
GPT3.5(COT)     & "\textcolor{blue}{Eventually} after many \textcolor{red}{events} of his \textcolor{blue}{anger} coming to the \textcolor{blue}{front}, we \textcolor{red}{fired} him.\\
Llama3          & "\textcolor{red}{Finally} after many \textcolor{red}{times} of his \textcolor{blue}{anger} coming to the \textcolor{blue}{front}, we \textcolor{red}{let} him \textcolor{red}{go}.\\
Llama3(COT)     & "\textcolor{red}{Finally} after many \textcolor{red}{times} of his \textcolor{blue}{anger} \textcolor{blue}{showing}, we \textcolor{red}{fired} him.\\
CoLLS(Llama3) & "\textcolor{red}{Finally} after many \textcolor{red}{events} of his \textcolor{blue}{anger} coming to the fore, we \textcolor{red}{fired} him.\\
CoLLS(GPT3.5) & "\textcolor{blue}{Eventually} after many \textcolor{red}{times} of his \textcolor{blue}{anger} coming to the fore, we \textcolor{red}{fired} him.\\\hline
\textbf{Inst.3}          & She was \textbf{located} and \textbf{fined} within two days of posting the \textbf{photograph}.\\\hline
LSBert          & She was \textcolor{blue}{located} and \textcolor{red}{charged} within two days of posting the \textcolor{red}{picture}.\\
GPT3.5          & She was \textcolor{red}{found} and \textcolor{blue}{fined} within two days of posting the \textcolor{red}{picture}.\\
GPT3.5(COT)     & She was \textcolor{red}{found} and \textcolor{blue}{fined} within two days of posting the \textcolor{blue}{photograph}.\\
Llama3          & She was \textcolor{red}{found} and \textcolor{red}{punished} within two days of \textcolor{blue}{sharing} the \textcolor{red}{picture}.\\
Llama3(COT)     & She was \textcolor{red}{found} and \textcolor{red}{punished} within two days of posting the \textcolor{red}{picture}.\\
CoLLS(Llama3) & She was \textcolor{red}{found} and \textcolor{blue}{cited} within two days of posting the \textcolor{red}{picture}.\\
CoLLS(GPT3.5) & She was \textcolor{red}{found} and \textcolor{red}{penalized} within two days of posting the \textcolor{red}{picture}.\\\hline
\textbf{Inst.4}          & Later, companies like Google, Yahoo, Tumblr and Vine tweeted with \textbf{hashtag} "\#LoveWins".\\\hline
LSBert          & Later, companies like Google, Yahoo, Tumblr and Vine \textcolor{blue}{called} with the \textcolor{blue}{``\#LoveWins ''}.\\
GPT3.5         & Later, companies like Google, Yahoo, Tumblr, and Vine \textcolor{blue}{posted} on Twitter with \\&\textcolor{blue}{hashtag} "\#LoveWins".\\
GPT3.5(COT)     & Later, companies like Google, Yahoo, Tumblr, and Vine \textcolor{blue}{posted} with \textcolor{blue}{hashtag} "\#LoveWins".\\
Llama3          & Later, companies like Google, Yahoo, Tumblr and Vine tweeted with \textcolor{blue}{hashtag} "\#LoveWins".\\
Llama3(COT)     & Later, companies like Google, Yahoo, Tumblr and Vine tweeted with \textcolor{blue}{hashtag} "\#LoveWins".\\
CoLLS(Llama3) & Later, companies like Google, Yahoo, Tumblr and Vine tweeted with \textcolor{red}{tag} "\#LoveWins".\\
CoLLS(GPT3.5) & Later, companies like Google, Yahoo, Tumblr and Vine tweeted with \textcolor{red}{tag} "\#LoveWins".\\\hline
\textbf{Inst.5}          & That said, I plan to \textbf{investigate} this question (among others) further {[}in{]} the next \textbf{couple} of years. \\\hline
LSBert          & That said, I plan to \textcolor{blue}{discuss} this question (among others) further {[}in{]} the next \textcolor{blue}{couple} of years.\\
GPT3.5          & That said, I \textcolor{blue}{intend} to \textcolor{red}{look into} this question (among others) more {[}in{]} the next \textcolor{blue}{couple} of years.\\
GPT3.5(COT)     & That said, I \textcolor{blue}{intend} to \textcolor{red}{look into} this question (among others) further {[}in{]} the next \textcolor{blue}{couple} of years. \\
Llama3          & That said, I plan to \textcolor{red}{look into} this question (among others) more {[}in{]} the next \textcolor{red}{few} years.\\
Llama3(COT)     & That said, I plan to \textcolor{red}{look into} this question (among others) more {[}in{]} the next \textcolor{red}{few} years.\\
CoLLS(Llama3) & That said, I plan to \textcolor{red}{examine} this question (among others) further {[}in{]} the next \textcolor{blue}{couple} of years.\\
CoLLS(GPT3.5) & That said, I plan to \textcolor{red}{explore} this \textcolor{blue}{query} (among others) further {[}in{]} the next \textcolor{red}{few} of years. \\\hline      
\end{tabular}
\caption{Five instances of different methods of simplification, with complex words in bold, correctly simplified words in red, and incorrect ones in blue.}
\label{tab:instances} 
\end{table*}
\end{document}